# ADAPTIVE DITHERING USING CURVED MARKOV-GAUSSIAN NOISE IN THE QUANTIZED DOMAIN FOR MAPPING SDR TO HDR IMAGE

Subhayan Mukherjee[1]  Guan-Ming Su[2]  Irene Cheng[1]

[1] University of Alberta, Edmonton, AB T6G 2R3, Canada
[2] Dolby Laboratories Inc, Sunnyvale, CA 94085, USA
{mukherje,locheng}@ualberta.ca  guanmingsu@ieee.org

**Abstract.** High Dynamic Range (HDR) imaging is gaining increased attention due to its realistic content, for not only regular displays but also smartphones. Before sufficient HDR content is distributed, HDR visualization still relies mostly on converting Standard Dynamic Range (SDR) content. SDR images are often quantized, or bit depth reduced, before SDR-to-HDR conversion, e.g. for video transmission. Quantization can easily lead to banding artefacts. In some computing and/or memory I/O limited environment, the traditional solution using spatial neighborhood information is not feasible. Our method includes noise generation (offline) and noise injection (online), and operates on pixels of the quantized image. We vary the magnitude and structure of the noise pattern adaptively based on the luma of the quantized pixel and the slope of the inverse-tone mapping function. Subjective user evaluations confirm the superior performance of our technique.

## 1  Introduction

High Dynamic Range (HDR) imaging technology with 12+ bits per color channel is becoming commonplace [1]. Traditional 8-bit Standard Dynamic Range (SDR) imaging only has a peak brightness of 100 nits and narrower color gamut compared to HDR (peak brightness of 1000+ nits and a wider color gamut). However, consumers can only make the most of this technology when more HDR content is widely available to the public. Current videos are mostly distributed at 8-bit depth. Although modern cameras can capture 12-/16-bit, videos are quantized to 8 bits SDR for compression and transmission. In order to watch SDR videos on HDR displays, the challenge is to up-convert the 8 bits content effectively for visualization, e.g. apply inverse tone-mapping operator [2-7]. HDR videos generated by these methods often suffer from false contours called banding/ringing artifacts, arising due to the Mach band effect [8,9]. 8-bit SDR video has a maximum of 256 code-words. Consequently, the output HDR video also has a maximum of 256 code-words when we deploy single-channel mapping. But in order to show a banding-free image on a 1000+ nits display, 12-bit (i.e., 4,096) code-words, is necessary [1]. Dithering techniques aim to mask banding by placing a combination of pixels with different colors in the neighborhood to perceptually mask banding artefacts [10].

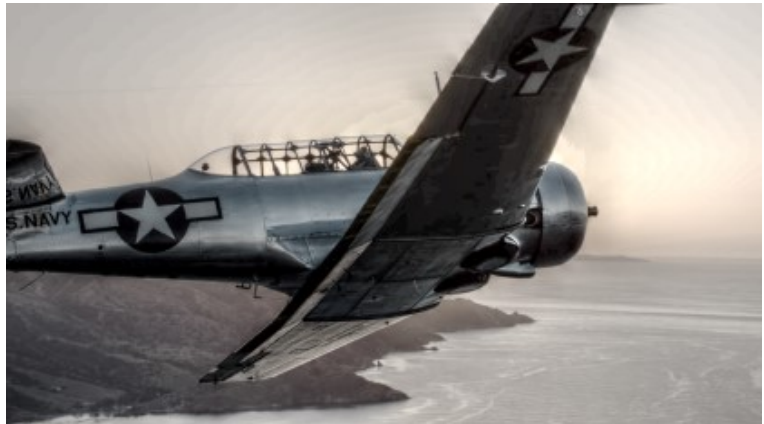

(a)

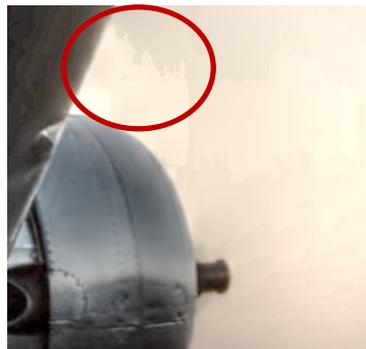 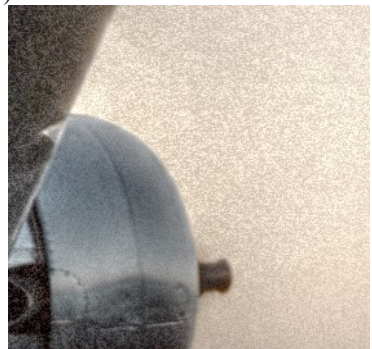

(b) (d)

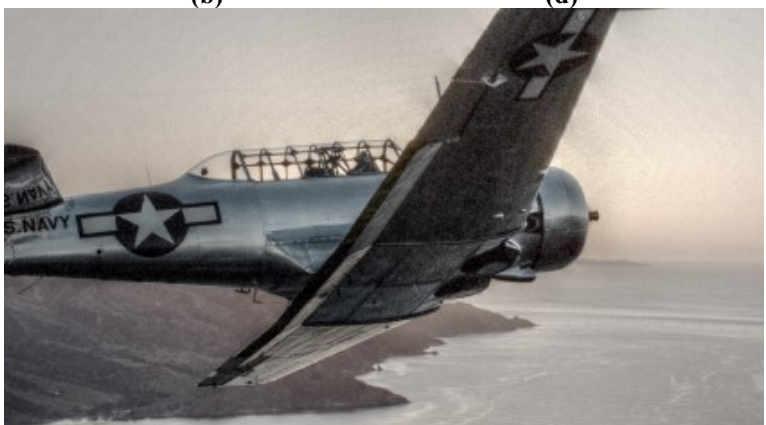

(c)

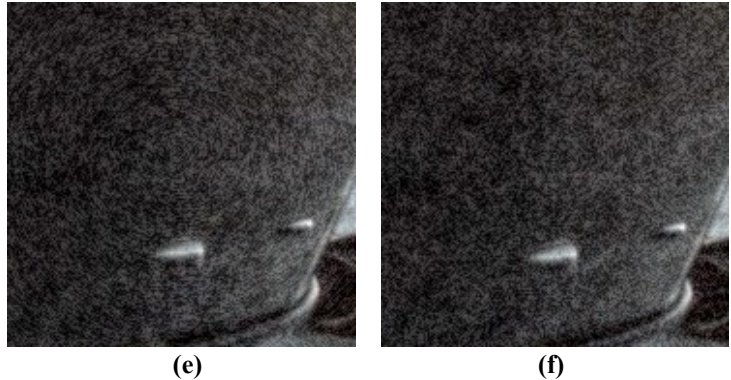

**(e)** **(f)**

**Fig. 1.** (a): A quantized Image and (b) a cropped region with banding. (c): Output of (a) from our method and (d) the corresponding cropped area. (e): Circular Noise, compared with (f): Curved Noise. This example shows that our curved noise method gives a better blending result and de-bands quantized images effectively. Note that (a-f) are tone-mapped [28] for illustration in printed form or on a SDR display (realistic visualization is only possible on a HDR display).

Traditional dithering methods can be adaptive or non-adaptive. The former uses both the pixel and its neighborhood information, whereas the latter uses only the pixel. Both approaches need to access the original high bit-depth and un-quantized pixel values [10-21]. However, many applications, e.g., video-on-demand, require content compression or data quantization before transmission. We proved mathematically that adding zero-mean noise can only remove banding from un-quantized data, but not from quantized signal. Furthermore, in some computing and/or memory I/O limited environment, the traditional solution using spatial neighborhood information is not feasible. Thus, our contribution lies in proposing a two-stage dithering method in the quantized domain, which is composed of two components: (1) the offline Curved Markov Gaussian noise pattern generation stage, and (2) the online luma/tone curve modulated noise injection stage.

### 1.1 Background

Since video transmission on mobiles is commonplace and HDR capable mobiles are increasingly popular in the consumer market, we consider the mobile GPU computing environment, where computation capacity is limited and accessing neighbor information is expensive in terms of memory I/O and processing time. Such environment limits the usage of filtering and prediction of false contour based on neighborhood information [22-27, 30]. Our method performs pixel-wise dithering on quantized images (display end), without the need to detect texture structure in the neighborhood. One application scenario, where only quantized images are available, is a video decoder which decodes 10-bit streams and outputs 8 bits signal (bit depth interface between two connected hardware chips are different) and another case is when 8-bit streams are received (e.g. 8-bit AVC/HEVC bit stream). In this context, the input quantized images undergo inverse tone-mapping to output HDR content of significantly higher bit-depth for display on the mobile device.

## 2  Proposed Computational Model

The proposed method is designed to process the computational expensive noise pattern generation offline, and to inject minimal noise to mask banding adaptively online. Fig. 1(a-d) illustrate our de-banding result, which keeps the image visually pleasant. Fig. 1(a, b) highlight how banding is visible especially in texture-less regions, e.g., sky, but is less visible in textured regions, e.g., airplane. Fig 1d shows the result of our noise injection, which is adaptively adjusted based on the degree of banding of the input. We simulate quantization by right shifting the original SDR input by 2 bits and then left-shifting it by 2 bits. Thus, the input is still 10 bits, but has very sparsely distributed code-words (banding). The output HDR image uses 16 bits per channel, but its code-words are normalized to the [0, 1) range. We consider a generic SDR-to-HDR conversion scheme where the luma channel is inverse tone-mapped using a look-up table. We label this look-up table BLUT (Backward Look-Up Table) as it is used to get back the HDR from the SDR image. But the chroma channels are processed using non-BLUT methods (i.e., we dither chroma channels without assuming any look-up tables). The range of luma intensities near the top and bottom of the code-word range are clipped, as shown in Fig.2a. This process of restricting the intensity range is commonly used in the Society of Motion Picture and Television Engineers (SMPTE) [29]. In the rest of this paper, we label the highest SDR intensity in the lower flat region as $Y_0$ and the lowest SDR intensity in the upper flat region as $Y_1$ respectively.

In our experiments, images are in the Y-$C_b$-$C_r$ color space, where Y is luma. $C_b$ and $C_r$ are the two chroma channels. We consider the inverse tone-mapping function as a one-to-one mapping between the SDR and HDR intensities in the Y channel. The advantage is to compute only once for each frame and store the result in BLUT, indexed by the SDR code-words. We consider that the original SDR uses 10 bits in each channel, and thus has 1024 code-words with integer values in the range [0, 1023]. Due to 8-bit quantization per channel, our method's input has only 256 code-words.

An example of BLUT is shown in Fig. 2a, where the horizontal axis has SDR code-words and the vertical axis has normalized HDR code-words. We found via testing that the smallest SDR intensity, with corresponding normalized HDR intensity greater than 0.625, can be taken as the starting intensity for highlight region. On a 4,000 nits display, this HDR intensity would be 3,000 nits. We denote the corresponding SDR intensity '$Y_h$', all individual SDR intensities $t$ and the set of all SDR intensities $T = \{t_0, t_1, t_2, \ldots, t_{1023}\}$. Now, let the lower flat region be $BLUT_{Down} = \{t < Y_0\}$, the low-lights & mid-tones region be $BLUT_{Mid} = \{Y_0 \leq t < Y_h\}$, the highlight region before reaching the upper flat region be $BLUT_{High} = \{Y_h \leq t < Y_1\}$, and the upper flat region be $BLUT_{Up} = \{t \geq Y_1\}$. In this way, we partition BLUT into four mutually exclusive regions: $T = \{BLUT_{Down} \cup BLUT_{Mid} \cup BLUT_{High} \cup BLUT_{Up}\}$.

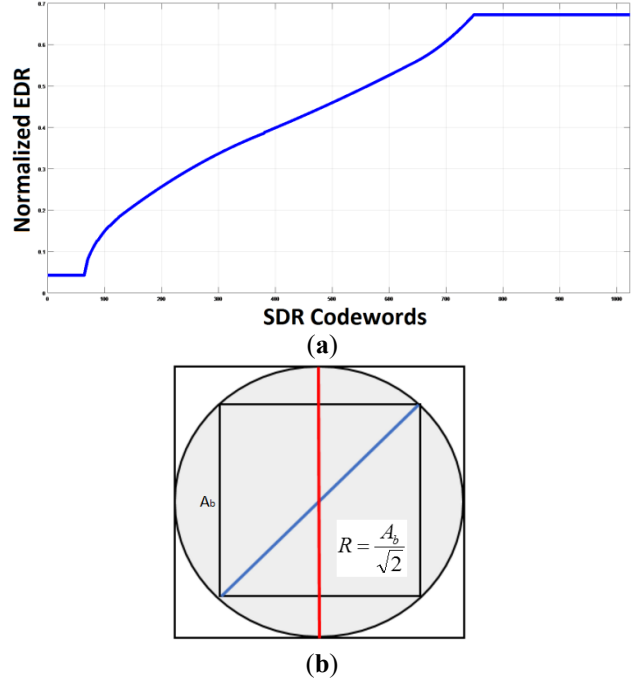

**Fig. 2.** (a) BLUT for the Y channel from a sunset scene, and (b) Extraction of the square block in a circular noise pattern.

### 2.1 Simple Two-State Markov Gaussian Noise Generation

Experimental analysis showed that when simple Gaussian noise is used for dithering, it masks the banding artefacts more effectively if the mean and standard deviation of the Gaussian distribution is increased, but at the same time it makes the image progressively noisier. Better results can be obtained by creating noise patterns, which has a global zero mean property, though locally the noise patterns have non-zero means to break banding. Using Markov chain to build state transition, where each state has non-zero mean noise can achieve this local non-zero mean.

As shown in Fig. 3, apply Markov Gaussian signal to generate noise for each pixel, we need to know the previous pixel's state. Based on the previous state and an intra-state probability $p$, we can stay in the same state or move to another state with an inter-state probability $1-p$. In each state, $s$, we can generate a Gaussian noise with mean $\mu_s$, variance $\sigma_s$, i.e. $(\mu_0, \sigma_0)$ for state $0$ and $(\mu_1, \sigma_1)$ for state $1$. We use the value $\mu_0 = 2, \sigma_0 = 1, \mu_1 = -2, \sigma_1 = 1$ in our implementation, as these values gave the best results. By comparing (a) and (b) in Fig. 4, we can see that a higher intra-state transition probability can generate longer texture, and have better ability to destroy the banding artefact, but make the image noisier.

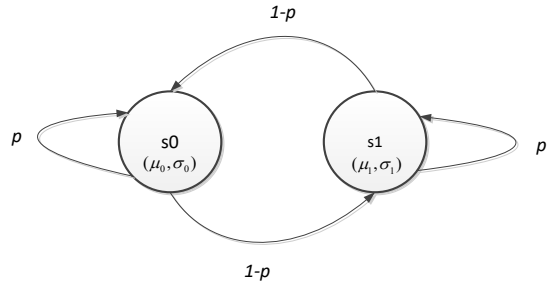

**Fig. 3.** Two-state Markov-Gaussian Noise generator

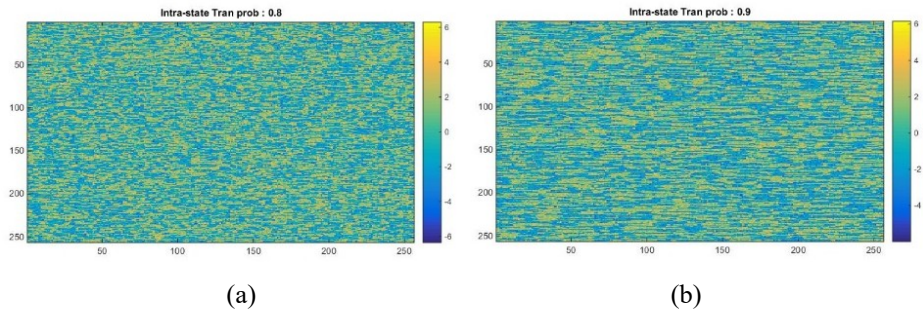

**Fig. 4.** Two-state Markov Gaussian Noise with intra-state transition probabilities (a) 0.8 and (b) 0.9. Note that higher transition probability has higher masking effect.

### 2.2  Limitation of Simple Markov Gaussian Noise

A limitation of the two-state Markov Gaussian noise is that it generates noise patterns ("stripes") in only two fixed directions: horizontal and vertical. But, such patterns look unnatural, and the effectiveness of de-banding depends on the angle at which the stripes meet the false contours.

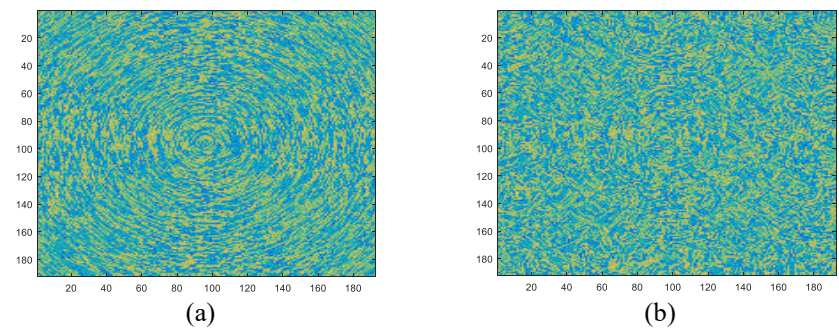

**Fig. 5.** Convert (a) Circular to (b) Curved Markov-Gaussian Noise; Block size = 200x200; Trans. Prob. = 0.815.

This motivated us to adopt the curved noise pattern instead of straight, in order to blend the patterns into the image content naturally. We first obtain concentric circles from the noise generator's output. In each circle, starting from the point on its circumference subtending the least angle $\theta$ (counter-clockwise) w.r.t the horizontal axis, we copy the output of the noise generator to that pixel. The length, '$L$' of the sequence, which the noise generator should generate to construct a square block of circular noise of size $A_b \times A_b$ is:

$$L = 2\pi \sum_{v=0}^{N_m} \left(\left(\frac{A_b}{\sqrt{2}}\right) - v\right) \qquad (1)$$

where $N_m = (R-1)$, and $R = 200/\sqrt{2} \sim 142$ (we use square blocks of each side length $A_b = 200$). This metric is illustrated in Fig. 2b. Radius of the biggest circle is R. Fig. 5a shows an example of circular noise pattern. Although Fig. 5 shows only one value of transition probability (0.815), our method in fact adaptively determines this value based on the slope of the BLUT in the current image.

### 2.3 Curved Markov Gaussian Noise from Circular Pattern

To generate a curved pattern from a circular pattern, we partition the square block of circular noise into four equal-sized *quadrants* or sub-blocks. We process each quadrant as an independent image and choose the same set of points at each quadrant as sites for Voronoi tessellation to generate irregular sized patches, called Voronoi cells (Fig. 6).

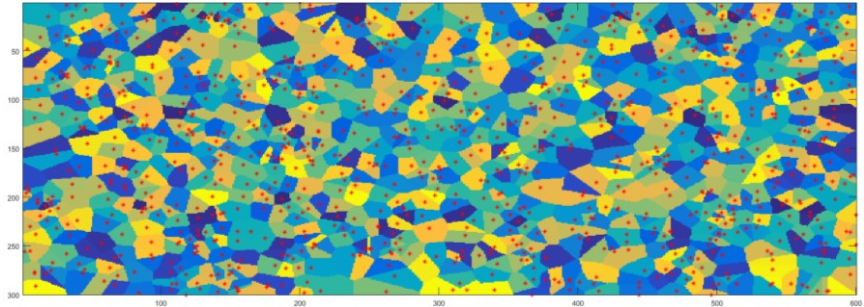

**Fig. 6.** Voronoi Tessellation of arbitrary matrix with adjacent cells shown in different colors; sites (*): red dots.

Since Voronoi tessellation is deterministic, we have one-to-one correspondence between the cells in all quadrants with respect to their shapes and sizes. Now, for each cell in each quadrant, we replace its content with that of any randomly selected cell from its corresponding set of four co-located cells across all quadrants. We integrate the quadrants to get the curved pattern block shown in Fig. 5b. The blocks are concatenated to get the noise matrix. Experiments showed that block size of 200x200 gave high-quality result, with a typical number of sites ($N_s$) = 300. Note that swapping square or rectangular blocks (instead of Voronoi cells) with each other generates blocky artefacts in curved noise, but randomizing site locations across frames ensures that noise patterns vary smoothly and the video looks more natural.

### 2.4 Noise Injection and Transition Probability

Noise patterns are generated offline only once and stored. When system boots up, they are loaded to memory. For noise injection, a higher transition probability gives better de-banding effect, but makes the image noisier. We can also increase the variance of noise to get rid of more banding, but making the image noisier at the same time. We define:

$$D = Q + s \cdot N_p \quad (2)$$

where '$s$' is the noise variance and $N_p$ is the matrix containing the generated noise pattern. $Q$ is the quantized image with banding artefacts. We inject noise to produce the dithered image $D$. Noise patterns, fetched from memory, have transition probabilities given by:

$P_T = \{0.545 + k * 0.045\}$ where $k = [0, 1, 2 \ldots 9]$.

An effective way to compute the transitional probability for any given intensity is to use the slope of BLUT, where the intensity is an indication of the potential degree of banding. The noise pattern for all pixels having that intensity are selected accordingly from memory, and blended into the image. Note that wherever the BLUT has a steep slope, adjacent SDR intensities are mapped to HDR intensities, which are far apart from each other. This increases the possibility of banding artefacts in the backward reshaped (HDR) image. Thus, in regions where the banding is more noticeable, we apply higher intra-state transition probability to inject more structured noise. The fine granularity of probabilities in $P_T$ ensures that the aggregated noise pattern applied to the entire quantized image does not have abrupt variations in adjacent regions.

## 3   Experiments and Analysis

In order to understand the visual impact of banding caused by the SDR-to-HDR conversion, we observed the Y, $C_b$, and $C_r$ channels in the output images on a 4,000 nits PULSAR HDR display. We created five video sequences[1] and selected 20 scenes from these sequences, with varying degrees of banding and luma intensities, and generated our test dataset of 120 frames (6 frames per scene). User evaluations were conducted by twelve subjects[2], who have computing science or engineering background. Images were randomly selected and shown in pairs on the PULSAR display with the light turned off in the room. Subjects were asked to decide the preferred image, or no preference, and enter the answer using a Google Form. If there was a preference, the subject was asked whether the preferred image was slightly better or much better than the other, and (optionally) whether the preferred image had lesser banding and/or lesser noise than the other. Based on the questionnaire, we could account for human errors when rating the subjective opinions. Five image pairs were shown in different order in each test as per Table 1. To avoid bias, the noise types were not revealed to the subjects. Note that, our proposed method is *adaptive* because it performs BLUT-slope modulated dithering. It does not need to use neighborhood information, which is needed in the traditional adaptive methods. The non-adaptive version of our curved noise technique uses fixed variance and fixed transition probability, but the adaptive version gives a better result.

---
[1] There are no public HDR datasets containing banding
[2] At least four subjects should be included, per T-REC P.910 ( https://www.itu.int/rec/T-REC-P.910/en )

**Table 1.** Image pairs displayed randomly to subjects

| Test-ID | ImageA | ImageB |
| --- | --- | --- |
| Test-1 | Simple Gaussian Noise | Proposed Adaptive Method Noise |
| Test-2 | Proposed Adaptive Method Noise | Non-adaptive Curved Noise Method |
| Test-3 | Low-Pass Filtered Gaussian Noise | Proposed Adaptive Method Noise |

The subjective opinion scores were assigned these values:

**1**. If a method is preferred by the subject as "much better than the other method" then it gets a score of +2

**2**. If a method is preferred by the subject as "slightly better than the other method" then it gets a score of +1

**3**. In case of "NO preference", both methods get a score of 0

Test results (Difference of Mean Opinion Scores, or DMOS) are shown in Table 2. Raw subjective scores collected for calculating these DMOS scores are available at: http://webdocs.cs.ualberta.ca/~mukherje/DMOS_RAW.xlsx

**Table 2.** DMOS scores from subjective test results

| Test-ID | Test-1 | Test-2 | Test-3 |
| --- | --- | --- | --- |
| DMOS | 1.067 | 0.717 | 1.033 |

The positive DMOS values of the adaptive method are promising. Subjective opinions showed that our method does not cause detrimental side-effect on image quality and the noise patterns generated by our dithering method are perceptually smooth across successive frames, which is important for video delivery.

### 3.1 Time Performance

We tested the serial implementation of our method in MATLAB running on Windows 10 64-bit, x64-based AMD FX-4350 4.2 GHz CPU, 16 GB RAM PC. We generated a set of Markov-Gaussian noise patterns with different intra-state probability. This of-fline step is computational intensive (11.79 seconds), but is done only once, stored and loaded to memory for the online noise injection stage. Adaptively selecting noise pattern, adjusting the noise magnitude and injecting to a 1920×1080 pixels frame takes 4.36 seconds. Faster performance can be achieved by converting the MATLAB code to a more efficient platform, e.g. C/C++.

## 4      Conclusion

We propose a pixel-based curved pattern dithering method for SDR-to-HDR conversion. In contrast to traditional methods, our adaptive approach works directly on quantized images without using spatial neighborhood information. Our method is more effective in some computing and/or memory I/O limited environment. Computational expensive noise generation step is performed offline. Subjective tests conducted on HDR display showed that our method de-bands significantly better compared to using other types of Gaussian noise with or without low-pass filter.